\documentclass{article} 
\usepackage{qwen}

\usepackage{microtype}
\usepackage{hyperref}
\usepackage{url}
\usepackage{booktabs}
\usepackage{amsmath}
\usepackage{lipsum}
\usepackage{graphicx}
\usepackage{natbib}
\usepackage{xcolor}
\usepackage{algorithm}
\usepackage{algorithmicx}
\usepackage{algpseudocode}
\usepackage{multirow}

\usepackage{lineno}

\definecolor{darkblue}{rgb}{0, 0, 0.5}
\hypersetup{colorlinks=true, citecolor=darkblue, linkcolor=darkblue, urlcolor=darkblue}

\title{SlimQwen: Exploring the Pruning and Distillation in Large MoE Model Pre-training}

\author{\textbf{Shengkun Tang}$^{1,\,2,\,\ddagger}$\thanks{Equal contribution. $^{\ddagger}$Work was done during the internship in Qwen Team. $^{\dag}$Co-Corresponding author. }, \textbf{Zekun Wang}$^{1,\,*}$, \textbf{Bo Zheng}$^{1,\,*}$, \textbf{Liangyu Wang}$^{1,\,3}$, \textbf{Rui Men}$^{1}$, \\ \textbf{Siqi Zhang$^{1}$, Xiulong Yuan$^{1}$, Zihan Qiu$^{1}$, Zhiqiang Shen$^{2,\,\dag}$, Dayiheng Liu$^{1,\,\dag}$}  \\
$^{1}$Qwen Team, Alibaba Inc., $^{2}$MBZUAI, $^{3}$KAUST 
}

\begin{document}


\maketitle

\begin{abstract}
Structured pruning and knowledge distillation (KD) are typical techniques for compressing large language models, but it remains unclear how they should be applied at pretraining scale, especially to recent mixture-of-experts (MoE) models. 
In this work, we systematically study MoE compression in large-scale pretraining, focusing on three key questions: whether pruning provides a better initialization than training from scratch, how expert compression choices affect the final model after continued training, and which training strategy is most effective. 
We have the following findings: First, across depth, width, and expert compression, pruning a pretrained MoE consistently outperforms training the target architecture from scratch under the same training budget.
Second, different one-shot expert compression methods converge to similar final performance after large-scale continual pretraining. Motivated by this, we introduce a simple partial-preservation expert merging strategy that improves downstream performance across most benchmarks.
Third, combining KD with the language modeling loss outperforms KD alone, particularly on knowledge-intensive tasks. We further propose multi-token prediction (MTP) distillation, which yields consistent gains.
Finally, given the same training tokens, progressive pruning schedules outperform one-shot compression, suggesting that gradual architecture transitions lead to better optimization trajectories.
Putting it all together, we compress Qwen3-Next-80A3B to a 23A2B model that retains competitive performance. These results offer practical guidance for efficient MoE compression at scale.

\end{abstract}
\section{Introduction}

Mixture-of-Experts (MoE)~\citep{shazeer2017outrageously} has become a dominant architecture for scaling large language models~\citep{jiang2024mixtral,qwen2.5,yang2025qwen3technicalreport,gemini2.5,qwen35blog}, but modern MoE LLMs remain expensive to pretrain and serve. Compressing a pretrained MoE into a smaller model that retains most of its capability at pretraining scale is therefore an important practical problem.

Structured pruning compresses models by removing entire architectural components (e.g., layers, attention heads, or experts) and delivers wall-clock speedups without specialized sparse kernels. Because pruning alone could degrade performance, knowledge distillation (KD) is commonly used to recover the loss by transferring knowledge from the teacher to the pruned student, and is widely believed to outperform continued pretraining with the standard language modeling (LM) objective. 
Despite extensive progress on dense models~\citep{muralidharan2024compactlanguagemodelspruning}, extending these compression paradigms to MoE models presents unique challenges.
Specifically, MoE models introduce an additional compression dimension: experts, which can be pruned or merged.
While recent studies~\citep{jaiswal2025findingfantasticexpertsmoes} thoroughly evaluate the one-shot performance of various expert compression methods, their efficacy following large-scale continual pretraining remains unexplored.

To bridge this gap, we revisit structured pruning and post-compression training for MoE LLMs by systematically investigating several practical questions: \textbf{(1) Initialization.} Does pruning a pretrained MoE model provide a stronger initialization than training an identical target architecture from scratch?
\textbf{(2) Compression Strategy.} How do different expert compression strategies impact final performance after extensive continual pretraining?
\textbf{(3) Training Recipe.} What is the optimal post-compression training recipe to facilitate performance recovery?

By exploring MoE-based LLM compression across depth, width, and experts via extensive continual pretraining, we present our key findings as follows: First, 
under the matched training tokens, pruning a pretrained MoE model to a target architecture provides a significantly better initialization than training from scratch, consistently improving both reasoning and generation performance.
Second, we conduct a comprehensive empirical analysis of expert compression and propose a partial-preservation strategy.
By comparing various pruning and merging criteria (e.g., routing frequency or scores, expert activations) under a 400B-token continual pretraining setting,
we find that the final performance differences among one-shot expert pruning or merging methods are marginal, with no single approach dominating. Motivated by this observation and the critical need to balance pretrained expert specialization against the consolidation of discarded experts, we propose a strategy that explicitly retains the top half of target experts intact while merging the less critical remainder into them. This prevents representation homogenization and consistently enhances downstream evaluation performance. 
Third, we demonstrate that hybridizing next-token knowledge distillation (NTP KD) with a standard language modeling (LM) loss, regulated by a linear decay schedule, yields superior recovery on knowledge-intensive benchmarks compared to pure KD. To further elevate the compacted model, we propose multi-token prediction~\citep{mtp} distillation (MTP KD). This paradigm extends the distillation objective beyond single tokens, fundamentally enhancing the backbone's training dynamics and representation quality, and improving the acceptance rate in multi-token speculative decoding. Finally, we study how to schedule pruning and distillation progressively when transitioning from a base architecture to a target architecture. Given a target configuration, we systematically compare direct one-stage compression against three progressive pruning schedules: depth-first, width-first, and joint. Across all configurations, progressive strategies consistently surpass one-shot pruning under an identical token budget. This confirms that staged capacity reduction provides a significantly smoother optimization trajectory for knowledge transfer.

Empirically, we demonstrate that our pruning and distillation recipe can compress the Qwen3-Next-80A3B~\citep{team2025qwen3next} to a 23A2B model (approximately $4\times$ compression) with competitive downstream performance after continual pretraining across a broad suite of evaluations, including MMLU variants, BBH, GSM8K, coding, and Chinese benchmarks. Overall, our results provide practical guidance for compute-efficient MoE compression at pre-training scale~\citep{qwen35blog}, clarifying (i) how structured pruning across depth/width/experts should be applied, (ii) how progressive schedules affect recovery, and (iii) which training objective is most effective during long post-compression training.
Our main contributions are:
\begin{itemize}
    \item We present a systematic study of large-scale MoE compression at pretraining scale, covering structured pruning initialization, expert compression, post-compression continual pretraining objectives, and progressive pruning schedules. We show that structured pruning provides a strong initialization, and that after large-scale continual pretraining, different one-shot expert pruning/merging methods yield similar final performance. We further propose a simple partial-preservation expert merging strategy that shows consistent improvement across benchmarks.
    \item We introduce the multi-token knowledge distillation that improves backbone model training and speculative decoding, and investigate different pretraining loss choices. Our experiments show that incorporating LM loss improves performance on knowledge-intensive benchmarks, while MTP KD yields consistent gains across the major benchmarks.
    \item We compare progressive pruning schedules and find that all progressive pruning strategies consistently outperform one-shot compression under the same final sparsity and total training tokens. Empirically, we compress Qwen3-Next-80A3B into a 23A2B model that achieves competitive performance across a wide range of benchmarks, including general reasoning, mathematics, and coding.
\end{itemize}
\section{Related Work}

\noindent \textbf{Structured Pruning in LLMs.} Structured pruning has been shown to be an effective technique to improve the model efficiency without specific hardware support. Considering MoE LLMs, there are three dimensions to prune: 1) width pruning such as hidden size and FFN intermediate size, 2) depth pruning, which removes whole transformer blocks by some metrics, and 3) expert pruning/merging including removing or merging a number of experts in MoE module. Some prior works such as ShearedLLaMA ~\citep{xia2024shearedllamaacceleratinglanguage} and SliceGPT ~\citep{ashkboos2024slicegptcompresslargelanguage} focus on width pruning in dense LLMs~\citep{muralidharan2024compactlanguagemodelspruning}. For depth pruning,  ShortGPT ~\citep{men2024shortgptlayerslargelanguage}, Laco ~\citep{yang2024lacolargelanguagemodel} and ShortenedLLaMA ~\citep{kim2024shortenedllamadepthpruning} all provide simple but effective methods to prune the depth of LLMs. \citet{cao2025condensedontjustprune} propose a method that merges large MoE layers into smaller dense layers. Moreover, M-SMoE ~\citep{li2024mergecompressdemystifyefficient} and REAP ~\citep{lasby2025reapexpertspruningprevails}
propose to merge the experts in the MoE modules to reduce the memory consumption while ~\citep{lu2024expertsequalefficientexpert} simply prune the redundant experts. In this work, we aim to achieve high compression ratio and combine depth/width pruning and expert pruning/merging. Furthermore, we propose a simple but effective expert merging technique, which improves the performance after post-compression training.

\noindent \textbf{Post-Compression Training for Recovery.} Since the model after structured pruning shows non-negligible performance degradation, post-compression training is generally required to recover the performance of the pruned model~\citep{ma2023llmpruner,wang2025cfsp}.  Minitron ~\citep{muralidharan2024compactlanguagemodelspruning} and Slim applies distillation to improve the performance of the pruned dense model while DarwinLM ~\citep{tang2025darwinlmevolutionarystructuredpruning} and SlimMoE ~\citep{li2025slimmoestructuredcompressionlarge} utilize conventional language modeling loss (LM loss) and KD respectively. However, Minitron is applicable only to non-MoE models, whereas DarwinLM and SlimMoE prune only the experts’ intermediate-layer dimensions within MoE modules. \citep{peng2024pretrainingdistillationlargelanguage} systematically studies pre-training distillation for LLMs, focusing on factors such as logits processing, loss selection, scaling law, and offline versus online teacher logits. In contrast, our work studies post-compression continual pretraining for large MoE models after structured pruning, with a focus on pruning initialization, expert pruning/merging, and training strategies after compression.

\begin{figure*}[t]
    \centering
    \includegraphics[width=1\linewidth]{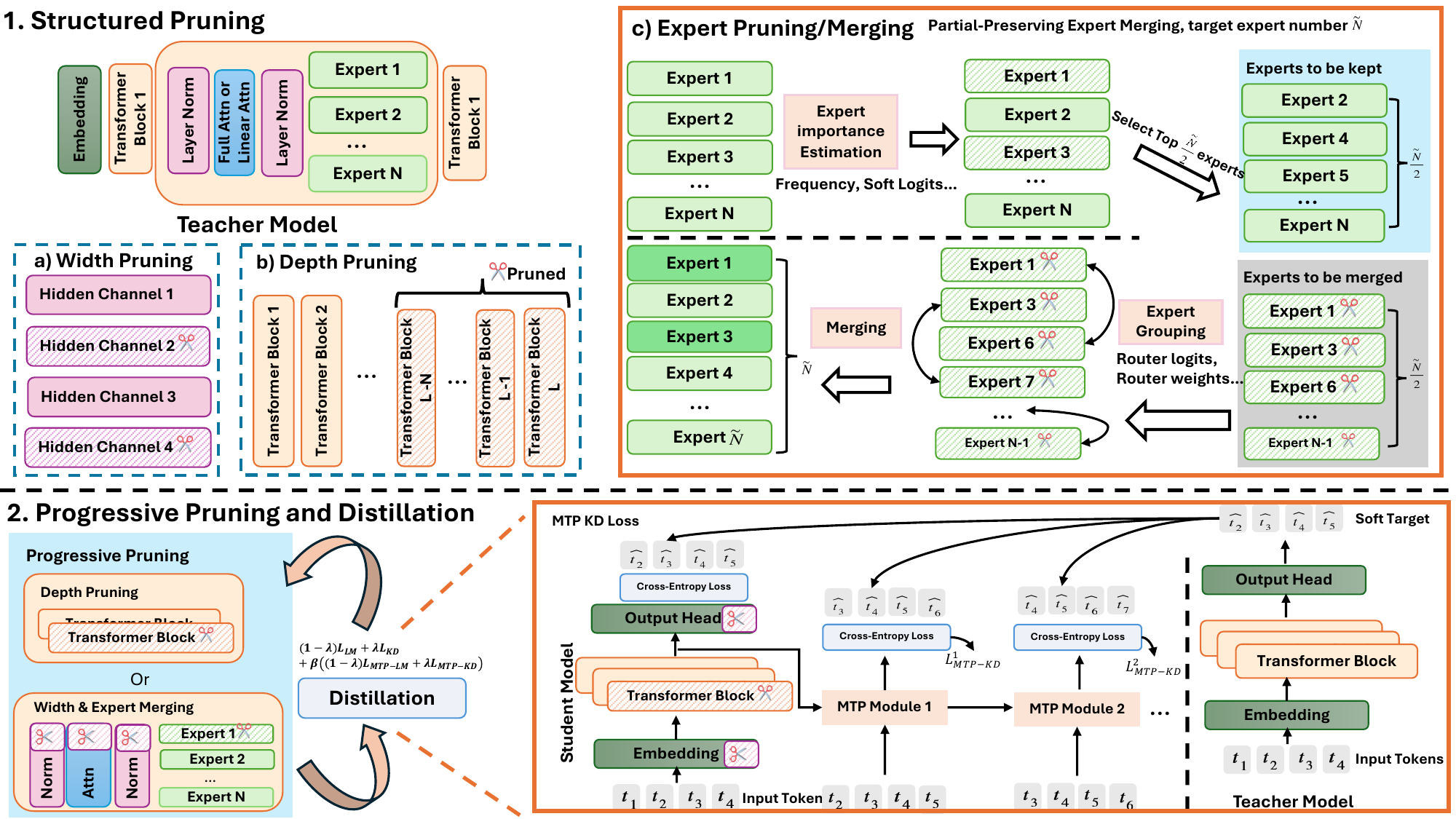}
    \caption{Overview of the SlimQwen.
    We first perform structured pruning on a teacher MoE model, including width pruning, depth pruning, and expert pruning/merging based on importance estimation and similarity, with a proposed partial-preservation strategy. We then adopt progressive pruning and distillation to gradually transform the teacher into the target architecture via staged pruning schedules (depth-first, width-first, or joint). Finally, we introduce a multi-token prediction (MTP) distillation, which extends standard next-token distillation by supervising multiple future tokens, improving training effectiveness. }
    \vspace{-1em}
    \label{fig:main_fig}
\end{figure*}

\section{Method}

\subsection{Background and Notation.}
Qwen3-Next~\citep{team2025qwen3next} is a hybrid-attention MoE-based model with $L$ layers, each block includes Gated DeltaNet ~\citep{yang2025gateddeltanetworksimproving} or Gated Attention modules ~\citep{qiu2025gatedattentionlargelanguage} with ratio $(L_{linear} : L_{full})$, MoE module with $N_e$ regular experts and $N_s$ shared experts, and RMSNorm modules. 

For the MoE module, given an input token $x\in\mathbb{R}^{1 \times d}$, we define $n$ experts in total, including
$n_{\mathrm{routed}}$ routed experts and $n_{\mathrm{shared}}$ shared experts
($n=n_{\mathrm{routed}}+n_{\mathrm{shared}}$). Each expert is a SwiGLU MLP:
\begin{equation}
\mathrm{Expert}(x)
=
(\mathrm{SiLU}(xW_{1e}) \odot (xW_{2e}))W_{3e},
\end{equation}
where $W_{1e},W_{2e}\in\mathbb{R}^{d\times d_{\mathrm{ff}}}$ and $W_{3e}\in\mathbb{R}^{d_{\mathrm{ff}}\times d}$.
The router produces top-$k$ gating scores over the routed experts:
$
z(x)=\mathrm{softmax}\,\!\big(\mathrm{TopK}(xW^{G},k)\big),\, W^{G}\in\mathbb{R}^{d\times n_{\mathrm{routed}}}.
$
In addition, we apply a separate shared gate $z_{\mathrm{s}}(x)=\sigma(xw_{\mathrm{sh}})\in\mathbb{R}^{n_{shared}}, \,w_{\mathrm{s}}\in\mathbb{R}^{d\times n_{shared}}$ for shared experts.
The MoE output is
\begin{equation}
\mathrm{MoE}(x)
=
\sum_{e=1}^{n_{\mathrm{routed}}} z_e(x) \mathrm{Expert}_e(x)
+
\sum_{s=1}^{n_{shared}}z_{\mathrm{s}}(x)\mathrm{Expert}_s(x).
\end{equation}
Qwen3-Next uses the RMSNorm~\citep{rmsnorm} normalizing function
\begin{align}
\text{RMSNorm}(X) = \frac{X}{\text{RMS}(X)} \odot \gamma, \quad  \text{RMS}(X)_i = \sqrt{\frac{1}{d} \sum_{j=1}^{d} X_{ij}^2 + \epsilon}
\end{align}
where $\text{RMS}(X) \in \mathbb{R}^{n \times 1}$ is the root mean square computed over the hidden dimension for each token, and $\gamma \in \mathbb{R}^{1 \times d}$ is the learnable scale parameter. The constant $\epsilon$ is added for numerical stability.
The details of Gated DeltaNet and Gated Attention can be found in the Appendix Sec. \ref{sec:arch}.

\subsection{MoE-based Model Compression}
In this work, we focus on exploring MoE-based Model compression across three dimensions: depth, width, and experts.
We introduce the details of strategy for each dimension below.

\noindent \textbf{Depth Pruning.} Considering a model with $L$ sequential layers $\{f_\ell\}_{\ell=1}^{L}$, we directly drop the last $N$ layers of an $L$-layer model \citep{sun2026cursedepthlargelanguage}\footnote{We provide the performance comparison and discussion of different depth pruning methods in Appendix Sec. \ref{sec:depth_pruning_comparison}. The last-layer pruning achieves better performance on both one-shot and continual pretraining settings.}:
\begin{equation}
\mathcal{L}_{\mathrm{keep}}=\{1,\dots,L-N\},\qquad \tilde{L}=L-N.
\end{equation}
In our experiments, we prune the last 25\% layers.

\noindent \textbf{Width Pruning.} 
For width pruning, we reduce the hidden dimension across the entire architecture, encompassing the hybrid attention, MoE, and normalization modules.
We estimate the importance of each hidden dimension using activation statistics computed on a sampled calibration dataset $\mathcal{D}$ from our training dataset. Let $Z\in\mathbb{R}^{B\times n\times m}$ denote the output activation of a module
for a batch size $B$, sequence length $n$, and hidden dimension $m$. We aggregate along the batch and sequence
dimensions using mean absolute activation: $\mathrm{Mean}(Z)
\;:=\;
\frac{1}{Bn}\sum_{b=1}^{B}\sum_{t=1}^{n}\big|Z_{b,t,:}\big|
\;\in\;\mathbb{R}^{m}.$
Let $Y=\mathrm{RMSNorm}(X)\in\mathbb{R}^{B\times n\times d}$ be the RMSNorm output . The hidden dimension importance are formulated as:
\begin{align}
I_{\mathrm{norm}}^{(k)}
=
\Big[\frac{\sum_{i=0}^L\mathrm{Mean}\big(\mathrm{RMSNorm}(X)\big)}{L}\Big]_k, \, k=1,\dots,d.
\end{align}
Given the target hidden size $d_t$, we retain the $d_t$ hidden dimensions with the highest importance scores.

\noindent \textbf{Expert Compression.}
Regarding expert compression, we compare various compression strategies, including pruning and merging. 
The initial step involves quantifying expert importance with various criteria. 
Given a set of calibration data, frequency-based criteria records the activated frequency while soft-logits method further weights frequency with the logits of router outputs for each expert.  We also consider the router-weighted expert output activation (REAP)~\citep{lasby-reap}. Formally, for each MoE layer, let there be $N$ routed experts $\mathcal{E}=\{E_1,\dots,E_N\}$ and a router
$R:\mathbb{R}^{d}\rightarrow\mathbb{R}^{N}$ that outputs routing logits
$z(x)=R(x)\in\mathbb{R}^{N}, x\in\mathbb{R}^{d}.$
For each token representation $x$, we select the top-$k$ experts
$\mathcal{A}(x)=\mathrm{TopK}(z(x),k)\subseteq\{1,\dots,N\}.$ let $E_j(x)$ be the expert output.
We can compute the frequency-based, soft-logits and REAP expert importance via:
\begin{align}
I_i^{\mathrm{Freq}} = \mathbb{E}_{x\sim\mathcal{C}}\Big[\mathbb{I}\big[i\in\mathcal{A}(x)\big]\Big], \qquad
I_i^{\mathrm{Soft}} = \mathbb{E}_{x\sim\mathcal{C}}\Big[\frac{\mathbb{I}[i\in\mathcal{A}(x)]\cdot z_i(x)}
{\sum_{j\in\mathcal{A}(x)}z_j(x)}\Big],
\\
I^{\mathrm{REAP}}_i = \frac{1}{|\mathcal{X}_i|}
\sum_{x\in\mathcal{X}_i}
z_i(x)\,\big\|E_i(x)\big\|_2,
\qquad i=1,\dots,N,
\end{align}
where $\mathbb{I}[\cdot]$ is the indicator function.
In practice, the expectation is computed by mean over all tokens in the calibration set.

For expert merging, we need to identify both the target clusters and the interpolation weights. 
We first quantify inter-expert similarities using router logits $z(x)$, router weights and output activation $E_j(x)$ among each expert. 
Given the above expert-importance scores, we preserve the highest-ranked experts. Each discarded expert is then merged into its nearest retained neighbor, using its importance score as the scaling factor. 
A central challenge in expert compression is striking an optimal balance between knowledge preservation and expert consolidation. Exclusively retaining top-ranked experts preserves highly salient knowledge but risks discarding experts that are individually less prominent yet functionally complementary. Conversely, constructing all target experts through aggressive merging can homogenize pretrained expert specialization, hindering performance recovery during continual pretraining. To navigate this trade-off, we propose a simple \textbf{partial-preservation merging strategy}: we retain half of the target experts intact, and construct the remainder by merging the discarded experts into selected merge bases. Formally, given a target number of retained experts $\tilde{N}<N$, we keep half target of experts with the largest importance scores:
$
\mathcal{S}_{\mathrm{keep}}
=
\operatorname*{arg\,topk}_{i\in\{1,\dots,N\}} I_i$
with $|\mathcal{S}_{\mathrm{keep}}|= \lfloor\tilde{N} // 2 \rfloor$
and the pruned expert index is $\mathcal{S}_{\mathrm{prune}}=\{1,\dots,N\}\setminus\mathcal{S}_{\mathrm{keep}}$. Finally, we select another $\tilde N/2$ experts from the remaining experts as merge bases, denoted by $\mathcal{S}_{\mathrm{base}}$. For each $i\in\mathcal{S}_{\mathrm{base}}$, we find its most similar partner
$
m(i)=\arg\max_{j\in\mathcal{S}_{\mathrm{merge}}}\mathrm{CosineSim}(i,j),$
and merge the two experts as
\begin{equation}
\tilde{E}_i
=
\frac{I_i}{I_i+I_{m(i)}}E_i
+
\frac{I_{m(i)}}{I_i+I_{m(i)}}E_{m(i)}.
\end{equation}
The final compressed expert set is composed of the preserved experts and the merged experts.
For both expert pruning and expert merging, we prune the corresponding router weight for continual pretraining. A detailed algorithm description can be found in Algorithm \ref{alg:half_preserving}.  We choose half of the target experts as a simple and symmetric design choice. Intuitively, preserving too few experts weakens parameter inheritance, whereas preserving too many leaves limited room for consolidation. Keeping roughly half provides a robust compromise in our evaluated setting. We discuss this more in Limitation section.

\subsection{Distillation Pretraining}

\noindent \textbf{MTP Distillation Loss.} We use Multi-Token Prediction (MTP) modules~\citep{mtp} to predict additional
future tokens. The MTP module consists of a embedding layer $\mathrm{Emb}(\cdot)$ and a output head $\mathrm{OutHead}(\cdot)$, which are shared with the backbone models. Moreover, a Transformer block $\mathrm{TRM}_k(\cdot)$ and a projection matrix
$M_k\in\mathbb{R}^{d\times 2d}$ are included in the MTP module.
For the $i$-th input token $t_i$, at prediction depth $k\in\{1,\dots,D\}$, we first combine the
representation of the $i$-th token at depth $k-1$, denoted by $h_i^{k-1}\in\mathbb{R}^{d}$, with the
embedding of the $(i+k)$-th token $\mathrm{Emb}(t_{i+k})\in\mathbb{R}^{d}$ via a linear projection:
\begin{equation}
h_i^{\prime k}
=
M_k\Big[\mathrm{RMSNorm}(h_i^{k-1});\,\mathrm{RMSNorm}\big(\mathrm{Emb}(t_{i+k})\big)\Big],
\label{eq:mtp_combine}
\end{equation}
where $[\cdot;\cdot]$ denotes concatenation. In particular, when $k=1$, $h_i^{0}$ refers to the token
representation produced by the main model.
The combined representation is then fed into the $k$-th Transformer block to produce the current-depth
representation: $h^{k}_{1:T-k}
=
\mathrm{TRM}_k\!\left(h^{\prime k}_{1:T-k}\right),
\label{eq:mtp_trm}
$
where $T$ is the sequence length and $1\!:\!T\!-\!k$ denotes slicing.
Finally, given $h_i^{k}$ as input, the shared output head computes the probability distribution for the
$k$-th additional prediction token:
$
p^{k}_{i+k}
=
\mathrm{OutHead}(h_i^{k})
\in\mathbb{R}^{V},
\label{eq:mtp_out}
$
where $V$ is the vocabulary size. The output head $\mathrm{OutHead}(\cdot)$ linearly maps $h_i^k$ to logits
and applies $\mathrm{Softmax}(\cdot)$ to obtain probabilities.

For each prediction depth $k\in\{1,\dots,D\}$, the $k$-th MTP module produces a student distribution
$p^{k}_{i+k}\in\mathbb{R}^{V}$ for position $i+k$. The MTP LM loss can be written as:
\begin{equation}
\mathcal{L}_{\mathrm{MTP\text{-}LM}}
=
\frac{1}{D}\sum_{k=1}^{D}
(
-\frac{1}{T-k}\sum_{i=1}^{T-k}\log p^{k}_{i+k}\!\left[t_{i+k}\right]
).
\end{equation}
Besides using ground-truth
one-hot labels, we distill from a teacher model that provides a soft target distribution
$q_{i+k}\in\mathbb{R}^{V}$ at the same position. We minimize the KL-divergence between teacher and student:
\begin{equation}
\mathcal{L}_{\mathrm{MTP\text{-}KD}}
=
-\frac{1}{D}\sum_{k=1}^{D}(
\frac{1}{T-k}\sum_{i=1}^{T-k}\sum_{v=1}^{V} q_{i+k}[v]\log p^{k}_{i+k}[v]
).
\end{equation}
where $T$ is the input sequence length and $V$ is the vocabulary size. 
Therefore, we train the model with four terms: (i) standard language modeling loss $\mathcal{L}_{\mathrm{LM}}$ and knowledge distillation loss $\mathcal{L}_{\mathrm{KD}}$ on the backbone output, MTP LM loss $\mathcal{L}_{\mathrm{MTP\text{-}LM}}$ and MTP distillation loss $\mathcal{L}_{\mathrm{MTP\text{-}KD}}$.
The total objective is
\begin{equation}
\mathcal{L}
=
(1-\lambda)\,\mathcal{L}_{\mathrm{LM}}
+
\lambda\,\mathcal{L}_{\mathrm{KD}}
+
\beta\,((1-\lambda)\mathcal{L}_{\mathrm{MTP\text{-}LM}} + \lambda \mathcal{L}_{\mathrm{MTP\text{-}KD}}).
\label{eq:loss}
\end{equation}
where $\lambda$ and $\beta$ are hyperparameters, which balance KD and LM loss, and backbone loss and MTP loss respectively.

\noindent \textbf{Progressive Pruning and Distillation.}
\label{sec:progressive}
Directly compressing a teacher model to a compact target architecture often induces substantial knowledge loss. To ensure a smoother transfer of pretrained capabilities, we explore three progressive, two-stage distillation schedules. Each schedule interleaves structural pruning with a fixed-token distillation phase, differing primarily in their reduction priorities for depth and width. \textbf{Depth-first} allocates half of the layer reduction to the first stage while maintaining the original width, leaving the remaining depth and the entire width reduction for the second stage. Conversely, \textbf{Width-first} executes half of the width reduction in the first stage while keeping the depth intact, completing the remaining width and the full depth reduction in the final stage. Finally, the \textbf{Joint} strategy simultaneously reduces both depth and width by half of their respective targets in the first stage, with the remaining halves pruned in the second stage to reach the final configuration.
Through this exploration, we aim to identify the optimal structural reduction trajectory that maximizes performance recovery during continual pretraining.
\section{Experiments}
\subsection{Experimental Setup}
\noindent \textbf{Base Model and Pruning Setup.} 
Unless otherwise noted, our experiments are conducted based on an 80A3B hybrid MoE-based model, which includes 48 transformer blocks with 12 full attention and 36 linear attention layers. Each full attention has 16 query heads and 2 key/value heads with 256 head dim. The gated attention ~\citep{qiu2025gatedattentionlargelanguage} is incorporated.
For the MoE layers, each module contains a total of 512 experts, with 10 routed experts and 1 shared expert activated per token.
The intermediate size is 512 and the hidden size is 2048. 
The model is trained with the multi-token prediction (MTP) module.
More architecture details can be found in Appendix Table \ref{tab:model_config}.  
For depth pruning, we remove 12 transformer blocks (3 full, 9 linear attention). In the remaining layers, we reduce the hidden size from 2048 to 1536. Additionally, we merge the 512 experts into 256 per MoE module and the compacted model activates only 8 routed experts with 1 shared expert per token.
We randomly use 1024 samples as calibration set to compute the importance metric.

\noindent \textbf{Training Settings.}  We evaluate our models under two training budgets: 120B and 400B high-quality, diverse tokens, with global batch sizes of 512 and 1024, respectively. The peak learning rate is set to 4e-4, decaying to 3e-5 via a cosine schedule with 2000 warmup steps. The distillation loss weight $\lambda$ decays linearly from 1 to 0.75, while the MTP distillation weight $\beta$ follows a cosine decay from 0.3 to 0.1. We train all models with the Maestro training framework~\citep{yuan2026accelerating}. We explain the detailed experiment settings in each section and details can be found in Appendix Table \ref{tab:training_hparams}. 

\noindent \textbf{Evaluation.}  We evaluate the few-shot performance of our models across a wide range of benchmarks. These include MMLU~\citep{hendrycks2021measuringmassivemultitasklanguage}, MMLU-Redux ~\cite{gema2025mmluredux} and MMLU-Pro ~\citep{wang2024mmluprorobustchallengingmultitask} for general knowledge; BBH ~\citep{suzgun2022challengingbigbenchtaskschainofthought} for reasoning; GSM-8K~\citep{cobbe2021trainingverifierssolvemath} for mathematics; EvalPlus~\citep{liu2023codegeneratedchatgptreally} for coding, C-Eval~\citep{huang2023cevalmultilevelmultidisciplinechinese} and CMMLU~\citep{li2024cmmlumeasuringmassivemultitask} for Chinese proficiency. We provide more evaluation in Appendix Sec. \ref{sec:more_eval}.

\subsection{Results} 
\begin{table*}[t]
    \centering
    \caption{The result comparison of models trained from scratch and initialized from pruned weights. The results show training from a pruned model brings benefits to the final model under the same training budget. $^\dagger$Here, the KD loss refers to the combined loss in Eq. \ref{eq:loss}.}
    \resizebox{\textwidth}{!}{
    \begin{tabular}{l|cccccccccc|c}
    \toprule
    \textbf{Method } &  \textbf{MMLU} & \textbf{MMLU-Pro} &\textbf{MMLU-Redux} & \textbf{BBH} & \textbf{GSM-8K}   & \textbf{EvalPlus} & \textbf{C-Eval} & \textbf{CMMLU}   & \textbf{Avg.}        \\ 
    
    \midrule
    Qwen3-Next-80A3B  &85.22 &62.86 &84.45 &85.12 &90.07 &74.12 &90.33 &89.27 &82.68  \\
    \hline
    Random Init + KD Loss$^\dagger$  &65.06 &34.54 &65.66 &56.01 &73.35 &58.67 &70.11 &69.85  &61.66   \\
    Pruned + LM Loss  &72.76 &48.24 &71.89 &64.94 &81.84 &67.05 &76.51 &76.51 &69.96 \\
    Pruned + KD Loss$^\dagger$  & \textbf{75.67} &\textbf{51.19} &\textbf{74.37} &\textbf{72.29} &\textbf{83.17} &\textbf{69.30} &\textbf{80.67} &\textbf{80.95} &\textbf{73.45} \\
    \bottomrule
    \end{tabular}
     }
     \vspace{-1em}
     
    \label{tab:prune_vs_random}
\end{table*}
\begin{figure*}[t]
    \centering
    \includegraphics[width=0.750\linewidth]{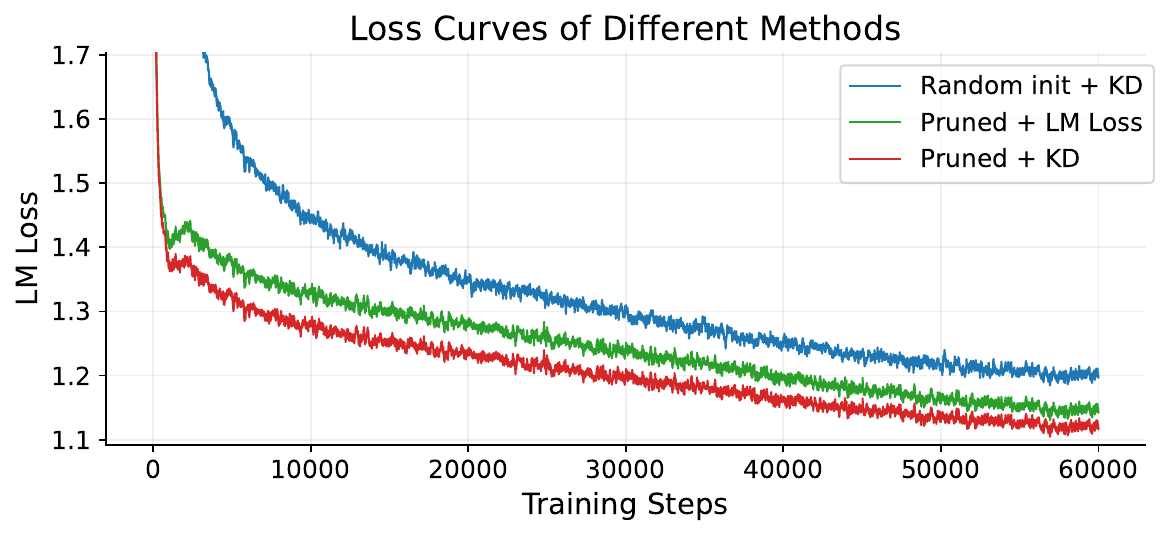}
    \caption{Training loss curves under different initialization and training objectives. Models initialized from pruned checkpoints converge faster and achieve lower LM loss than random initialization. Incorporating KD further improves optimization, with Pruned + KD consistently achieving the lowest loss, followed by Pruned + LM Loss, demonstrating the advantage of pruning-based initialization and distillation for efficient and effective training. }\
    \label{fig:training_curve}
    \vspace{-2em}
\end{figure*}
\noindent \textbf{Q1: Does pruning provide a better initialization for MoE in large-scale pretraining?} We first validate the effectiveness of training from a pruned MoE model in pretraining. As detailed in Table~\ref{tab:prune_vs_random}, both setups are trained for 120B tokens using knowledge distillation (KD) from the Qwen3-Next teacher. Compared to random initialization, the pruned model demonstrates striking superiority, achieving an average score of 73.45 against 61.66 (+11.79 points).
This consistent improvement spans diverse domains, including knowledge (MMLU), math (GSM-8K), and coding (EvalPlus). Remarkably, the pruned architecture recovers 86.5\% of the teacher’s performance (73.45 vs. 82.68) despite being 3.4$\times$ smaller, suggesting that structured pruning successfully preserves task-critical weights to form an informative starting point.
Furthermore, the training trajectories (Figure~\ref{fig:training_curve}) corroborate these findings: pruned initialization yields considerably faster convergence and lower language modeling (LM) loss than random initialization, with the combined "Pruned + KD" recipe achieving the lowest loss.

\noindent \textbf{Q2: How do different expert compression strategies impact final performance?} To evaluate various expert compression strategies, we compress a 24A2B MoE model to a 6A1B architecture and continually pretrain for 400B tokens.
As summarized in Table~\ref{tab:expert_pruning_compare}, \textit{no single one-shot pruning or merging method establishes consistent superiority across all downstream tasks,} even if some models show higher performance on certain benchmarks (e.g. frequency-based router logits grouping method achieves 60.17 on BBH). A possible explanation is that one-shot expert compression (coarse-grained pruning or merging) methods are unable to preserve the performance of all benchmarks consistently.
Furthermore, partial expert preservation during the merging experts yields consistent improvements across major benchmarks, including MMLU, MMLU-Pro, and GSM8K.

\noindent \textbf{Q3: What constitutes an effective training recipe for compressed MoEs?}
\begin{table*}[t]
    \centering
    \caption{Performance comparison of models with and without the partial-preservation expert merging strategy during expert pruning, and across different pruning and merging methods after continual pretraining. The results demonstrate that (1) the partial-preservation expert merging strategy leads to performance gains on major benchmarks, and (2) no single model exhibits uniformly superior performance across all evaluation tasks. }
    \resizebox{\textwidth}{!}{
    \begin{tabular}{c|c|c|c|cccccccccc|c}
    \toprule
    \textbf{Prune/Merging} &\textbf{Imp. Metric} &\textbf{Group Method} &\textbf{Preserve}   &  \textbf{MMLU} & \textbf{MMLU-Pro} &\textbf{MMLU-Redux} & \textbf{BBH} & \textbf{GSM-8K}   & \textbf{EvalPlus} & \textbf{C-Eval} & \textbf{CMMLU}         \\ 
    \midrule
    Expert Merging &Soft Logits  &Router Weights &No  &69.05 &42.62 &68.47 &59.12 &71.08 &50.35 &71.08 &72.20 \\
    Expert Merging &Soft Logits  &Router Weights &Yes  &\textbf{69.28} &\textbf{44.05} &67.64 &\textbf{59.81} &\textbf{74.18} &48.00 &\textbf{71.15} &\textbf{72.73}  \\
    \hline
    Expert Pruning &Soft Logits&- &-  &68.74 &\textbf{43.23} &\textbf{69.11} &58.97 &\textbf{74.30} &51.69 &71.67 &72.26  \\
    Expert Pruning &REAP&- &-  &69.11 &42.76 &67.57 &59.00 &73.69 &53.59 &71.67 &71.92  \\
    Expert Merging &Frequency  &Router Logits &Yes  &68.92 &42.14 &68.29 &\textbf{60.17} &72.82 &48.91 &70.26 &72.76  \\
    
    Expert Merging &Soft Logits  &Router Logits &Yes  &68.73 &42.35 &68.03 &59.88 &72.86 &52.11 &70.04 &71.85  \\
    Expert Merging &Soft Logits  &Expert Vector &Yes  &68.88 &42.47 &68.32 &59.00 &70.74 &49.69 &71.38 &72.23  \\
    Expert Merging &REAP  &Router Logits &Yes  &\textbf{69.74} &42.75 &67.46 &57.77 &73.69 &50.95 &\textbf{72.57} &72.22  \\
    Expert Merging &REAP  &Expert Vector &Yes &69.26 &42.93 &67.78 &59.45 &73.73 &\textbf{55.29} &71.45 &\textbf{72.89}  \\
    \bottomrule
    \end{tabular}
     }
     
    \label{tab:expert_pruning_compare}
\end{table*}
To establish an effective post-compression continual training recipe under pre-training setting, we evaluate various loss configurations on a 23A2B model pruned from Qwen3-Next-80A3B and trained for 120B tokens (Table~\ref{tab:training_loss}).
Our analysis reveals several findings:
Combining next-token prediction knowledge distillation (NTP KD) with a standard language modeling (LM) loss outperforms pure distillation, particularly on knowledge-intensive benchmarks such as MMLU (from 74.16 to 74.93) and MMLU-Pro (from 50.97 to 51.44).
Furthermore, ablations demonstrate that integrating multi-token prediction knowledge distillation (MTP KD) into either pure NTP KD or a comprehensive joint objective (NTP KD + LM + MTP loss) improves the performance on several knowledge-intensive benchmarks.
Beyond backbone quality, MTP KD yields substantial efficiency gains for speculative decoding across both pretraining and supervised fine-tuning (SFT), as shown in Table \ref{tab:mtp_acceptance_rate}. We report the results on benchmarks including, HumanEval, GSM8K, WMT22\citep{kocmi-etal-2022-findings} for pretraining stage, RepoQA\citep{liu2024repoqaevaluatinglongcontext}, MTBench\citep{chen2026mtbenchmultimodaltimeseries} and SpecBench\citep{xia-etal-2024-unlocking} for SFT stage.  
The results show that MTP KD consistently improves the multi-token acceptance rate from \textit{acc\_1} to \textit{acc\_4} on all benchmarks. A notable pattern is that the gains from MTP KD are often larger for longer accepted token sequence. This suggests that MTP KD is particularly helpful for improving the efficiency of multi-token generation, making the drafted tokens more likely to be accepted by the verifier model during speculative decoding. Overall, these results indicate MTP KD not only improves backbone training quality, but also brings practical benefits for speculative decoding.

\begin{table*}[t]
    \centering
    \caption{The benchmark performance comparison of different training losses. All models are pruned from Qwen3-Next-80A3B to 23A2B and trained on 120B tokens. Adding LM loss improves knowledge benchmarks (e.g., MMLU, MMLU-Pro), while incorporating MTP KD yields consistent gains, with the full objective achieving strong performance on several major benchmarks. NTP KD: Next-Token prediction knowledge distillation.}
    \resizebox{\textwidth}{!}{
    \begin{tabular}{l|cccccccccc}
    \toprule
    \textbf{Method } &  \textbf{MMLU} & \textbf{MMLU-Pro} &\textbf{MMLU-Redux} & \textbf{BBH} & \textbf{GSM-8K}   & \textbf{EvalPlus} & \textbf{C-Eval} & \textbf{CMMLU}          \\ 
    
    \midrule
    NTP KD  &74.16 &50.97 &\textbf{75.85} &71.63 &\textbf{84.27} &67.32 &80.00 &80.24   \\
    NTP KD + LM Loss  &74.93 &51.44 &74.69 &\textbf{73.00} &82.98 &66.07 &79.93 &80.31  \\
    NTP KD + MTP KD  &\textbf{75.13} &\textbf{51.94} &74.33 &71.93 &82.34 &\textbf{69.32}  &\textbf{80.82} &\textbf{80.64}  \\
    \hline
    NTP KD + LM Loss + MTP Loss  &75.29 &51.16 
 &\textbf{75.09} &72.07 &83.02 &68.43 &79.78 &80.67\\
    NTP KD + LM Loss + MTP Loss  + MTP KD  &\textbf{75.67} &\textbf{51.19} &74.37 &\textbf{72.29} &\textbf{83.17} &\textbf{69.30} &\textbf{80.67} &\textbf{80.95} \\

    \bottomrule
    \end{tabular}
     }
    \label{tab:training_loss}
\end{table*}

\begin{table}[t]
\centering
\caption{MTP generation acceptance rate (\%) by speculative decoding across pretraining and supervised-finetuning (SFT) stages. The results show that on both pretraining and SFT stages, compared with MTP Loss, MTP KD improves the multi-token generation acceptance rate consistently on most benchmarks.}
\label{tab:mtp_acceptance_rate}

\resizebox{\textwidth}{!}{%
\begin{tabular}{llccccccccccccccc}
\toprule
\multirow{2}{*}{\textbf{Stage}} & \multirow{2}{*}{\textbf{Loss}} & \multicolumn{5}{c}{\textbf{HumanEval}} & \multicolumn{5}{c}{\textbf{GSM8K}} & \multicolumn{5}{c}{\textbf{WMT22}} \\
\cmidrule(lr){3-7} \cmidrule(lr){8-12} \cmidrule(lr){13-17}
 &  & acc\_0 & acc\_1 & acc\_2 & acc\_3 & acc\_4 & acc\_0 & acc\_1 & acc\_2 & acc\_3 & acc\_4 & acc\_0 & acc\_1 & acc\_2 & acc\_3 & acc\_4 \\
\midrule
\multirow{2}{*}{Pretrain} & MTP Loss & \textbf{95.37} & 56.31 & 24.35 & 9.79 & 4.09 & \textbf{95.90} & 57.62 & 23.64 & 8.02 & 2.37 & \textbf{81.44} & 43.97 & 18.86 & 5.99 & 1.66 \\
 & MTP KD   & 94.77 & \textbf{68.60} & \textbf{37.06} & \textbf{17.36} & \textbf{8.24} & 95.50 & \textbf{75.18} & \textbf{45.67} & \textbf{22.43} & \textbf{10.37} & 81.29 & \textbf{49.04} & \textbf{24.56} & \textbf{10.03} & \textbf{3.97} \\
\bottomrule
\end{tabular}%
}
\resizebox{\textwidth}{!}{%
\begin{tabular}{llccccccccccccccc}
\toprule
\multirow{2}{*}{\textbf{Stage}} & \multirow{2}{*}{\textbf{Loss}} & \multicolumn{5}{c}{\textbf{RepoQA}} & \multicolumn{5}{c}{\textbf{MTBench}} & \multicolumn{5}{c}{\textbf{SpecBench}} \\
\cmidrule(lr){3-7} \cmidrule(lr){8-12} \cmidrule(lr){13-17}
 &  & acc\_0 & acc\_1 & acc\_2 & acc\_3 & acc\_4 & acc\_0 & acc\_1 & acc\_2 & acc\_3 & acc\_4 & acc\_0 & acc\_1 & acc\_2 & acc\_3 & acc\_4 \\
\midrule
\multirow{2}{*}{SFT} & MTP Loss & 96.02 & 64.68 & 29.36 & 11.23 & 3.91 & 87.85 & 57.40 & 28.72 & 12.46 & 4.93 & 87.61 & 55.58 & 27.73 & 12.02 & 4.60 \\
 & MTP KD   & \textbf{96.17} & \textbf{69.49} & \textbf{35.94} & \textbf{15.67} & \textbf{6.59} & \textbf{88.55} & \textbf{61.30} & \textbf{33.10} & \textbf{16.03} & \textbf{7.04} & \textbf{87.97} & \textbf{59.85} & \textbf{32.21} & \textbf{15.22} & \textbf{6.56} \\
\bottomrule
\end{tabular}%
}
\end{table}

\begin{table*}[!htbp]
    \centering
    \caption{The result comparison of one-shot and progressive pruning and distillation. All models are pruned to 23A2B. One-shot pruning trains directly on 400B tokens, while progressive pruning uses a two-stage strategy (40B + 360B). Progressive methods consistently outperform one-shot pruning on most benchmarks, highlighting the benefits of gradual pruning during pretraining.}
    \resizebox{\textwidth}{!}{
    \begin{tabular}{l|c|cccccccccc}
    \toprule
    \textbf{Method } &\textbf{Tokens} &  \textbf{MMLU} & \textbf{MMLU-Pro} &\textbf{MMLU-Redux} & \textbf{BBH} & \textbf{GSM-8K}   & \textbf{EvalPlus} & \textbf{C-Eval} & \textbf{CMMLU}          \\ 
    
    \midrule
    One-stage  &400B &75.86 &52.97 & 75.41 &73.97 &85.22 &70.07 &\textbf{83.87} &82.26 \\
    Joint  & 40B + 360B &76.30 &53.12 &76.93   &71.40 &\textbf{86.05} &70.58 & 83.57 & 82.62\\
    Width-first  & 40B + 360B & 77.14 & 52.80 &77.07 &\textbf{75.22} &84.00 &71.40 &82.01 &82.76 \\
    Depth-first (\textbf{SlimQwen})  & 40B + 360B & \textbf{77.39} &\textbf{53.22} &\textbf{78.01} &70.70 &85.82 &69.08 &82.97 &\textbf{83.01} \\
    \bottomrule
    \end{tabular}
     }
    \label{tab:progressive}
\end{table*}

\noindent \textbf{Progressive pruning and distillation.} Building upon the one-shot strategy, we further explore the efficacy of progressive pruning and distillation.
Given the final target architectural configuration, we progressively prune the base model using three strategies: depth-first, width-first, and joint pruning, each conducted in two stages as described in Section ~\ref{sec:progressive}. In the first stage, the intermediate pruned model is trained with 40B tokens. We then further prune it to the final target configuration and continue training on the remaining 360B tokens.
The results are shown in Table ~\ref{tab:progressive}.
Overall, progressive pruning and distillation consistently outperform one-stage pruning trained directly on 400B tokens, demonstrating the benefit of gradual model compression during continual pretraining. In particular, MMLU improves from 75.86 (one-stage) to 77.39 (depth-first) and 77.14 (width-first), while MMLU-Redux shows substantial improvements, from 75.41 to 78.01 and 77.07. These findings confirm that a progressive trajectory mitigates information loss and better transfers pretrained knowledge. We further provide the results for more fine-grained stage schedules in Appendix Sec. \ref{sec:app_progressive}. However, However, more fine-grained stage partitions do not provide additional benchmark performance gains. Given the superior overall performance, we officially designate the depth-first progressive model as \textbf{SlimQwen}.

\section{Conclusion}
In this paper, we explore the pruning and distillation in MoE model pretraining. We show that structured pruning, even at high compression ratios, provides a strong initialization for continual pretraining, while different expert pruning and merging metrics exhibit only minor differences after large-scale pretraining. We further propose a simple partial-preservation expert merging strategy and demonstrate consistent performance improvements across major benchmarks. For distillation, we investigate the effectiveness of progressive pruning and distillation, as well as the role of LM loss as a complementary training objective. We propose a novel multi-token prediction (MTP) distillation objective for pretraining, demonstrating consistent performance gains across major benchmarks.



\bibliography{colm2026_conference}
\bibliographystyle{colm2026_conference}

\appendix
\newpage

\section{Appendix}
\subsection{Architecture Details}
\label{sec:arch}
We provide the architecture details of the original teacher model and the pruned student models in Table \ref{tab:moe_configs}. Specifically, for the Gated Attention, given input hidden states $X\in\mathbb{R}^{n\times d}$, where $d$ is the model hidden size and $h_q$ is the number of query head, Gated Attention can be formulated as:
\begin{equation}
\mathrm{GatedAttn}(X)=
\mathrm{Concat}\,\!\big({\mathrm{head}}_{1}\odot g_i(X),\ldots,{\mathrm{head}}_{h_q}\odot g_{{h_q}}(X)\big)W_O, \,g_i(X)=\sigma(Xw_g^{(i)})\in\mathbb{R}^{n\times 1}, 
\end{equation}
where $W_O\in\mathbb{R}^{(h_q d_{\mathrm{head}})\times d}$ is the output matrix, $\sigma(\cdot)$ indicates the sigmoid function $\sigma(z)=\frac{1}{1+e^{-z}}$ and $w_g^{(i)}\in\mathbb{R}^{d\times 1}$ is a learnable gate weight.
The attention head is computed by scaled dot-product attention:
$
\mathrm{head}_{i}
=
\mathrm{Attn}\!\big(Q^{(i)},K^{(m(i))},V^{(m(i))}\big),
\mathrm{Attn}(Q,K,V)
=
\mathrm{softmax}(\frac{QK^\top}{\sqrt{d_{\mathrm{head}}}})V.
$
The per-head query, key and value projections are
$
Q^{(i)} = XW_Q^{(i)},
K^{(j)} = XW_K^{(j)},
V^{(j)} = XW_V^{(j)}
$
with learnable parameters $W_Q^{(i)},W_K^{(j)},W_V^{(j)} \in \mathbb{R}^{d\times d_{\mathrm{head}}}.$
We use Grouped-Query Attention (GQA) with $h_q$ query heads and $h_{kv}$ key/value heads.
For the Gated DeltaNet, we maintain a linear state matrix
$
S_t \in \mathbb{R}^{d_v \times d_k},
q_t \in \mathbb{R}^{d_k},
k_t \in \mathbb{R}^{d_k},
v_t \in \mathbb{R}^{d_v}.
$
The gated delta rule updates the state as
\begin{equation}
S_t
\;=\;
S_{t-1}\Big(\alpha_t\big(I-\beta_t k_t k_t^\top\big)\Big)
\;+\;
\beta_t\, v_t k_t^\top,
\qquad
\alpha_t\in(0,1),\ \beta_t\in(0,1).
\end{equation}
The token-mixing output is read out by $y_t \;=\; S_t q_t \in \mathbb{R}^{d_v}.$ We can map it back to the model dimension $d$:
$
Y_t = y_t W_{\mathrm{out}}\in\mathbb{R}^{d}, \,
W_{\mathrm{out}}\in\mathbb{R}^{d_v\times d}.
$
In our implementation, $d_k$ corresponds to the $Q/K$ hidden size and $d_v$ corresponds to the $V$ hidden size.  
\begin{table}[h]
\centering
\small
\setlength{\tabcolsep}{8pt}
\caption{Model configurations and parameter counts for different MoE variants.}
\resizebox{\textwidth}{!}{
\begin{tabular}{lcccccccccccc}
\toprule
\multirow{2}{*}{\textbf{Model}} 
& \multirow{2}{*}{$d_{\mathrm{model}}$}
& \multicolumn{5}{c}{\textbf{Self Attn}}
& \multicolumn{4}{c}{\textbf{MoE}}
& \multirow{2}{*}{$n_{\mathrm{MTP}}$}
 \\
\cmidrule(lr){3-7}  \cmidrule(lr){8-11}
& 
& $n_{\mathrm{qhead}}$ 
& $n_{\mathrm{kvhead}}$ 
& $d_{\mathrm{head}}$
& $n_{\mathrm{layer}}$
& Attn-Gate
& $d_{\mathrm{expert}}$
& $n_{\mathrm{expert}}$
& $n_{\mathrm{shared\mbox{-}expert}}$
& top-k
& & \\
\midrule
80B-A3B        & 2048 & 16 &2 &256 & 12 &Yes & 512 & 512 & 1 & 10 &1  \\
SlimQwen-23A2B  & 1536 & 16 &2 &256 & 8  &Yes & 512 & 256 & 1 & 8  &1  \\
\midrule
23B-A2B        & 2048 & 16 &2 &256  & 7 &Yes  & 512 & 256 & 1 & 8 &1  \\
SlimQwen-6A1B   & 1280 & 16 &2 &256 & 5 &Yes & 512 & 128 & 1 & 8 &1 \\
\bottomrule
\end{tabular}
\label{tab:model_config}
}

\resizebox{\textwidth}{!}{
\begin{tabular}{lcccccccccc}
\multirow{2}{*}{\textbf{Model}} 
& \multicolumn{8}{c}{\textbf{Linear Attn}}
& \multirow{2}{*}{\shortstack{\# Total\\Param}}
& \multirow{2}{*}{\shortstack{\# Act.\\Param}} \\
\cmidrule(lr){2-9} 
& $n_{\mathrm{vhead}}$
& $n_{\mathrm{qkhead}}$
& $d_{\mathrm{vhead}}$
& $d_{\mathrm{qkhead}}$
& $d_{\mathrm{conv}}$
& $d_{\mathrm{inner}}$
& $n_{\mathrm{layer}}$
& Attn-Gate 
& & \\
\midrule
80B-A3B         & 32 &16 &128 &128 &4 &4096 & 36 &No  & 80B & 3.8B \\
SlimQwen-23A2B   & 32 &16 &128 &128 &4 &4096 &24 &No  & 23B & 2.0B \\
\midrule
23B-A2B         & 32 &16 &128 &128 &4 &4096 & 21 &No  & 23B & 2B \\
SlimQwen-6A1B   & 32 &16 &128 &128 &4 &2560 &15 &No  & 6B & 1B \\
\bottomrule
\end{tabular}
}
\label{tab:moe_configs}
\end{table}

\subsection{Training Hyperparameters}
We provide the detailed pretraining hyperparameters in Table \ref{tab:training_hparams}.

\begin{table}[h]
\centering
\small
\setlength{\tabcolsep}{10pt}
\renewcommand{\arraystretch}{1.15}
\caption{Training hyperparameters for the 120B-token and 400B-token settings. The two settings differ only in the global batch size.}
\resizebox{0.6\textwidth}{!}{
\begin{tabular}{lcc}
\toprule
\textbf{Hyperparameter} & \textbf{120B Setting} & \textbf{400B Setting} \\
\midrule
Training tokens & 120B & 400B \\
Global batch size & 512 & 1024 \\
Learning rate & \multicolumn{2}{c}{$4\times10^{-4}$} \\
LR schedule & \multicolumn{2}{c}{Cosine decay} \\
Minimum learning rate & \multicolumn{2}{c}{$3\times10^{-5}$} \\
Warmup steps & \multicolumn{2}{c}{2000} \\
KD loss weight $\lambda$ & \multicolumn{2}{c}{Linear decay from 1.0 to 0.75} \\
MTP distillation weight $\beta$ & \multicolumn{2}{c}{Cosine decay from 0.3 to 0.1} \\
Calibration samples & \multicolumn{2}{c}{1024} \\
Training platform & \multicolumn{2}{c}{Alibaba Cloud} \\
\bottomrule
\end{tabular}
}
\label{tab:training_hparams}
\end{table}

\subsection{Implementation Detail}
Our codebase is built upon Megatron-LM. Following Qwen3 MoE models \citep{yang2025qwen3technicalreport}, we apply the global-batch load balancing loss \citep{qiu2025demonsdetailimplementingload} for MoE. The calibration data is sampled from the pretraining data. For progressive pruning distillation, we train all models using a single-stage learning rate decay schedule, such that the second stage starts from the learning rate reached at the final step of the first stage. We use the AdamW optimizer and apply the default hyperparameter settings for the optimizer. For speculative decoding, we use the MTP module as the draft model and backbone model as the verification model, and report \textit{acc\_0} as the acceptance rate of generating one token with the MTP module, \textit{acc\_1} as that of generating two tokens, and so on.  We also provide the pseudo-code of the partial-preservation expert merging strategy, as shown in Algorithm ~\ref{alg:half_preserving}.
\begin{algorithm}[h]
\caption{Partial-preservation Expert Merging Strategy}
\label{alg:half_preserving}
\begin{algorithmic}[1]
\Require Experts $\{E_i\}_{i=1}^N$, target expert number $\tilde{N}$, importance scores $\{S_i\}_{i=1}^N$
\Ensure Compressed experts $\tilde{\mathcal{E}}$

\State $S_{\mathrm{keep}} \gets \operatorname{arg\,topk}_{i\in\{1,\dots,N\}} S_i$, where $|S_{\mathrm{keep}}|=\lfloor \tilde{N}/2 \rfloor$
\State Select $S_{\mathrm{base}} \subset \{1,\dots,N\}\setminus S_{\mathrm{keep}}$ such that $|S_{\mathrm{base}}|=\tilde{N}-|S_{\mathrm{keep}}|$
\ForAll{$i \in \{1,\dots,N\}\setminus(S_{\mathrm{keep}}\cup S_{\mathrm{base}})$}
    \State $m(i)\gets \arg\max_{j\in S_{\mathrm{base}}}\mathrm{CosineSim}(i,j)$
    \State Assign $i$ to the merge group of $m(i)$
\EndFor
\ForAll{$j\in S_{\mathrm{base}}$}
    \State Merge all experts assigned to $j$ into $\tilde{E}_j$
\EndFor
\State \Return $\tilde{\mathcal{E}}=\{E_i:i\in S_{\mathrm{keep}}\}\cup\{\tilde{E}_j:j\in S_{\mathrm{base}}\}$
\end{algorithmic}

\end{algorithm}

\subsection{Comparison of Different Depth Pruning Methods}
\label{sec:depth_pruning_comparison}
We compare different depth pruning methods including activation similarity based method and pruning the last several layers directly, as shown in Table \ref{tab:depth_prune_methods}. Formally, Let $h_\ell\in\mathbb{R}^{n\times d}$ be the activation of layer $\ell$ and
$a_\ell=\frac{1}{n}\sum_{t=1}^{n}h_{\ell,t}\in\mathbb{R}^{d}$ be its token-mean pooled vector.
We compute adjacent-layer cosine similarity:
\begin{equation}
c_\ell=\frac{\langle a_{\ell-1},a_\ell\rangle}{\|a_{\ell-1}\|_2\|a_\ell\|_2},\qquad \ell=2,\dots,L.
\end{equation}
Let $\ell^\star=\arg\max_{\ell\in\{2,\dots,L\}} c_\ell$ be the starting index, and prune a contiguous
chunk of $N$ layers:
$
\mathcal{S}_{\mathrm{prune}}=\{\ell^\star,\ell^\star+1,\dots,\ell^\star+N-1\}$  and 
$\mathcal{S}_{\mathrm{keep}}=\{1,\dots,L\}\setminus\mathcal{S}_{\mathrm{prune}}.
$ 

We conduct the experiments on a pretrained 15A3B teacher model with 24 layers. We utilize the same 1024 calibration dataset discussed above to compute the layer activation and prune 4 layers in the one-shot setting. The activation-based pruning method tends to prune the middle layers. The results from the table indicate that directly pruning the last 4 layers causes only minor degradation (e.g from 75.62 to 73.86 on MMLU) while activation-based method shows substantially larger performance drops (e.g from 75.62 to 41.95 on MMLU). The results also align with the observation from \citep{sun2026cursedepthlargelanguage}. After 120B tokens of post-compression KD, last-layer pruning still recovers better performance than the activation-based method. One interesting phenomenon from the table is that the performance of the model trained with 120B tokens is worse than the one-shot counterpart on benchmarks such as MMLU and CMMLU. A possible explanation is that the one-shot performance is already close to that of the teacher model, leaving a relatively small knowledge gap to recover.

\begin{table*}[t]
    \centering
    \caption{The results comparison of different depth pruning methods in one-shot and continued pretraining settings. In the one-shot setting, pruning the last layer results in only minor degradation on benchmarks such as MMLU, whereas activation-based methods lead to substantially larger performance drops. After post-compression KD with 120B token, last-layer pruning still recovers better performance than the activation-based method.}
    \resizebox{0.7\textwidth}{!}{
    \begin{tabular}{l|cccc}
    \toprule
    \textbf{Method } &  \textbf{MMLU} & \textbf{CMMLU} &\textbf{CEval} & \textbf{GSM8K}        \\ 
    
    \midrule
    15A2B Teacher Model  &75.62 &81.35 &82.08 &82.41    \\
    \hline
    Activation Similarity  &41.95 &43.41 &42.28 &\textbf{11.22}  \\
    Last Layer Pruning  & \textbf{73.86} &\textbf{80.3} &\textbf{79.96} &2.05 \\
    \hline
    Activation Similarity + 120B tokens  &69.57 &74.32 &75.69 &73.84  \\
    Last Layer Pruning + 120B tokens  & \textbf{73.02} &\textbf{78.08} &\textbf{78.07} &\textbf{77.86} \\
    \bottomrule
    \end{tabular}
     }
     \vspace{-1em}
     
    \label{tab:depth_prune_methods}
\end{table*}

\subsection{Results of Progressive Pruning and Distillation with More Stages}
\label{sec:app_progressive}
We provide the results of progressive pruning and distillation with more fine-grained stages, as shown in Table \ref{tab:appendix_progressive}. There are two types of three-stage settings: depth-first and width-first. In the depth-first setting, we first prune half of the layers to be removed and train the model for 20B tokens, then prune the remaining half and train for another 20B tokens. Finally, in the third stage, we prune the width and continue training for 360B tokens. The width-first setting follows the same procedure in the reverse order. The results show that the three-stage settings achieve performance comparable to the two-stage setup. Although some three-stage variants perform better on individual benchmarks, the overall results remain similar. This suggests that a two-stage progressive pruning strategy is already sufficient in our setting.
\begin{table*}[!h]
    \centering
    \caption{The result comparison of one-shot and progressive pruning and distillation with 3 stages. More fine-grained stage partitions do not yield additional performance gains compared with the two-stage setup.}
    \resizebox{\textwidth}{!}{
    \begin{tabular}{l|c|cccccccccc}
    \toprule
    \textbf{Method } &\textbf{Tokens} &  \textbf{MMLU} & \textbf{MMLU-Pro} &\textbf{MMLU-Redux} & \textbf{BBH} & \textbf{GSM-8K}   & \textbf{EvalPlus} & \textbf{C-Eval} & \textbf{CMMLU}          \\ 
    
    \midrule
    One-stage  &400B &75.86 &52.97 & 75.41 &73.97 &85.22 &70.07 &83.87 &82.26 \\
    Width-first 3 Stages & 20B + 20B + 360B & 76.46 &52.18 &77.18 &73.44 &84.08 &68.32 &83.72 &82.70 \\
    Depth-first 3 Stages & 20B + 20B + 360B   &77.29 &52.63 &77.29 &73.37 &84.15 &71.12 &83.75 &82.65 \\
    
    Depth-first (\textbf{SlimQwen})  & 40B + 360B & 77.39 &53.22 &78.01 &70.70 &85.82 &69.08 &82.97 &83.01 \\
    \bottomrule
    \end{tabular}
     }
     \vspace{-1em}
    \label{tab:appendix_progressive}
\end{table*}

\subsection{Evaluation on More Benchmarks}
\label{sec:more_eval}
Due to the page limit, we provide evaluation results on more benchmarks of our experiments in this section. We further add CEval \citep{huang2023cevalmultilevelmultidisciplinechinese} for Chinese knowledge,  SuperGPQA \citep{pteam2025supergpqascalingllmevaluation} for general knowledge, KOR-Bench \citep{ma2025korbenchbenchmarkinglanguagemodels} and ICLEval \citep{chen2024iclevalevaluatingincontextlearning} for reasoning and in-context learning ability, MBPP \citep{austin2021programsynthesislargelanguage} for coding tasks, MMMLU \citep{hendrycks2021measuringmassivemultitasklanguage} and IncludeBase \citep{romanou2024include} for multilingual knowledge, and Mgsm \citep{shi2022language} for multilingual math ability. We provide the results comparison of models trained from scratch and initialized from pruned weights on these benchmarks in Table \ref{tab:prune_vs_random_appendix}.
\begin{table*}[!h]
    \centering
    \caption{More benchmark results comparison of models trained from scratch and initialized from pruned weights. }
    \resizebox{\textwidth}{!}{
    \begin{tabular}{l|cccccccccc|c}
    \toprule
    \textbf{Method } &  \textbf{CEval} & \textbf{SuperGPQA} &\textbf{KOR-Bench} & \textbf{ICLEval} & \textbf{MBPP}   & \textbf{MMMLU} & \textbf{IncludeBase} & \textbf{Mgsm}   & \textbf{Avg.}        \\ 
    
    \midrule
    Random Init + KD Loss  &70.11 &21.16 &33.36 &52.45 &57.00 &50.90 &42.83 &39.98  &45.97   \\
    Pruned + LM Loss  &76.51 &27.16 &39.52 &62.79 &63.2 &62.39 &55.96 &61.85 &56.17 \\
    Pruned + KD Loss  & \textbf{80.67} &\textbf{29.22} &\textbf{40.80} &\textbf{65.88} &\textbf{67.40} &\textbf{66.40} &\textbf{58.88} &\textbf{64.47} &\textbf{59.21} \\
    \bottomrule
    \end{tabular}
     }
     \vspace{-1em}
     
    \label{tab:prune_vs_random_appendix}
\end{table*}
\begin{table*}[!h]
\centering
\caption{Speedup and memory analysis of SlimQwen and the original model.}
\label{tab:efficientqwen_speed}
\renewcommand{\arraystretch}{1.15}
\setlength{\tabcolsep}{6pt}
\resizebox{\textwidth}{!}{
\begin{tabular}{l|c|cc|cc}
\toprule
\textbf{Model} & \textbf{Peak Memory (GB)} & \multicolumn{2}{c|}{\textbf{HF backend}} & \multicolumn{2}{c}{\textbf{vLLM backend}} \\
\cline{3-6}
& & \textbf{Prefill Latency (s)} & \textbf{Decoding Throughput (Tok/s)} & \textbf{Prefill Latency (s)} & \textbf{Decoding Throughput (Tok/s)} \\
\hline
Qwen3-Next-80A3B    & 156.56 & 0.99 & 4.05 & 0.08 & 142.58 \\
SlimQwen-23A2B & 43.30  & 0.44 & 6.55 & 0.06 & 210.87 \\
\bottomrule
\end{tabular}}
\end{table*}

\subsection{Efficiency Analysis}

We provide the efficiency analysis of SlimQwen and the original teacher model, as shown in Table \ref{tab:efficientqwen_speed}. The prompt length is 128 and the generation length is limited at 128. The process is executed for 10 times with 3 warmup runs and the results are computed as the average results. We provide the results with HuggingFace and vLLM as the inference backend, respectively. The models are run on the same two GPUs with a tensor parallel size of 2. The peak memory is monitored with data type as bfloat16. We can observe that SlimQwen obtains better speedup on both prefilling and decoding. More importantly, as a small-size model, SlimQwen can be deployed on single GPU with 80GB memory, which can further boost the efficiency since no parallel strategies are requires such as Tensor-Parallel (TP) or Pipeline-Parallel (PP).

\end{document}